\documentclass[runningheads]{llncs}

\usepackage[T1]{fontenc}

\usepackage{amsmath, amsfonts}

\usepackage{orcidlink}

\usepackage{graphicx}
\usepackage{subcaption}

\usepackage[inline]{enumitem}
\setlist*[itemize]{labelindent=10pt, itemindent=0pt, leftmargin=*}

\usepackage{booktabs}
\usepackage{multirow}
\usepackage{tabularx}
\usepackage{pgfplots}
\pgfplotsset{compat=1.18}
\usepgfplotslibrary{statistics}

\usepackage[nameinlink]{cleveref}

\usepackage{csquotes}
\usepackage{textcomp}

\usepackage{balance}

\begin{document}

% ===============================================
% Title & acknowledgements
% ===============================================
\title{Explainable AI to Improve Machine Learning Reliability for Industrial Cyber-Physical Systems}
\titlerunning{XAI to Improve ML Reliability for Industrial CPS}

% ===============================================
% Authors
% ===============================================
\author{
Annemarie {Jutte}\inst{1,2}\textsuperscript{*}\orcidID{0009-0004-9322-0674} \and
Uraz {Odyurt}\inst{3}\textsuperscript{*}\orcidID{0000-0003-1094-0234}}

% \author{Redacted authors segment ...}

% ============================================================
% Short author
% ============================================================
\authorrunning{A. Jutte and U. Odyurt}

% ============================================================
% Affiliations
% ============================================================
\institute{
Ambient Intelligence Group, Saxion University of Applied Sciences, Enschede, The Netherlands \and
Department of Computer Science, University of Twente, Enschede, The Netherlands
\email{a.m.p.jutte@saxion.nl}\\
\and
Faculty of Engineering Technology, University of Twente, Enschede, The Netherlands
\email{u.odyurt@utwente.nl}}

% ===============================================
% Make title segment
% ===============================================
\maketitle

% ===============================================
% Author footnote
% ===============================================
\renewcommand{\thefootnote}{\fnsymbol{footnote}}
\footnotetext[1]{These authors contributed equally to this work.}

% ###############################################
% Abstract
% ###############################################
\begin{abstract}
Industrial Cyber-Physical Systems (CPS) are sensitive infrastructure from both safety and economics perspectives, making their reliability critically important. Machine Learning (ML), specifically deep learning, is increasingly integrated in industrial CPS, but the inherent complexity of ML models results in non-transparent operation. Rigorous evaluation is needed to prevent models from exhibiting unexpected behaviour on future, unseen data. Explainable AI (XAI) can be used to uncover model reasoning, allowing a more extensive analysis of behaviour. We apply XAI to improve predictive performance of ML models intended for an industrial CPS use-case. We analyse the effects of components from time-series data decomposition on model predictions using SHAP values. Through this method, we observe evidence on the lack of sufficient contextual information during model training. By increasing the window size of data instances, informed by the XAI findings for this use-case, we are able to improve model performance.

\end{abstract}

% ###############################################
% Text body
% ###############################################
% ###############################################
% Start of file - body.tex
% ###############################################

% ===============================================
% Section
% ===============================================
\section{Introduction}
\label{sec:introduction}
Cyber-Physical Systems (CPS), particularly industrial CPS, are considered as sensitive infrastructure, both in terms of safety and economic security. Solutions incorporating Machine Learning (ML) are often utilised in the lifecycle of industrial CPS, e.g., for health monitoring, and therefore must be highly reliable.

Considering the ML model component of CPS, evaluation methods beyond common ML model performance metrics, i.e., prediction accuracy, F1 score, etc., are in demand. While these metrics indicate how well a model performs on the test data, they do not reveal \emph{why} it performs well or demonstrate its ability to generalise to \emph{unseen data}. Consequentially, spurious correlations used by the model, which are present in both training and test data, but are not actual causal relationships, cannot be detected. Understanding the relations a model relies on is important for robust evaluation and alignment with domain knowledge.

Deep neural networks, often employed as ML solutions, are considered black boxes due to their complexity~\cite{Adadi:2018:PIBB}. Evaluating such models' decision rationale is non-trivial. Explainable AI (XAI) techniques such as SHAP~\cite{Lundberg:2017:UAIM} and LIME~\cite{Ribeiro:2016:WSTY} can be used to provide insights into the reasoning of AI-based systems~\cite{Adadi:2018:PIBB}.

XAI can be used during model development to inform the design process, and during model deployment to support human decision making. By understanding model behaviour, training hyperparameters can be adjusted based on empirical deductions, rather than design-space search methodologies. \emph{We focus on the use of XAI for informed model development}. It must be mentioned that the high computational cost is the downside. Considering the context of our use-case, ensuring reliability of CPS is far more important than the computational cost.

Application of XAI methods to practical CPS use-cases remains limited. The context of industrial CPS is especially interesting as time-series data, commonly associated with such systems, encapsulates complex behaviour. As a consequence, explanations regarding model-reasoning based on time-series might be less intuitive to understand than reasoning, for example, image data. One option is to explain model reasoning in terms of high-level human-interpretable concepts, specifically components from time-series decomposition~\cite{Jutte:2025:CTSA}.

\paragraph*{Proposed solution}
We generate and incorporate SHAP values for the fine-tuning of the ML model module, a Convolutional Neural Network (CNN), in a Fault Detection and Identification (FDI) solution. We specifically leverage decomposition of the time-series pseudo-signals\footnote{A time-series is the collection of data points ordered in time, which can be considered as signal sampling.}
and well-argued adjustments to data formatting based on model's response to signal components. Given the component scoring, upon consistent reliance on a particular component, better representation of that component per data instance could have a positive effect. For instance, a wider data instance could better reflect certain component states, or a narrower data instance could result in better focus on abrupt variations.

\paragraph*{Contribution}
We provide:
\begin{itemize}
    \item The application of SHAP to an industrial CPS use-case. Specifically, the application of concept-based SHAP utilising a custom decomposition of time-series signals.
    \item A demonstration of how model performance can be improved during model development through XAI-informed decision making.
    \item Publicly available data~\cite{Data} and code~\cite{Code} for reproduction.
\end{itemize}

The use-case serves as a demonstration of achievable improvements. The approach can be extended to further improve a targeted ML model. Following this introduction, this paper is organised in Background, Model improvement, Implementation, and Outcome, \Cref{sec:background,sec:model_improvement,sec:implementation,sec:outcome}, respectively. After a concise related work in \Cref{sec:related_work}, concluding remarks are given in \Cref{sec:conclusion}.

% ===============================================
% Section
% ===============================================
\section{Background}
\label{sec:background}
We discuss the relevance of \emph{execution phases} in industrial CPS to ML models and provide an overview of XAI methods.

\subsection{Industrial CPS data structures}
Execution phases are segments of time-series monitoring data collected from industrial CPS, reflecting the behaviour of the system under scrutiny per individual task during an execution~\cite{Odyurt:2021:PPFT}. As industrial CPS operates primarily in sequential and repetitive machine cycles, the concept of phases is an effective method of data compartmentalisation. As such, data related to each task or sub-task can be analysed individually and in isolation. Phases covering tasks and respective sub-tasks form a hierarchical relation. Compartmentalised and repetitive phase data are most suitable for ML dataset instances.

\subsection{Explainable AI}
XAI approaches generally fall into two categories: explaining black box models or developing inherently interpretable \enquote*{white box} models. While white box models offer transparency, due to the restrictions on their architecture, these often provide lower performance than black box models~\cite{Adadi:2018:PIBB}. Post-hoc and model-agnostic explainability methods can be applied to any predictive model.

SHAP (SHapley Additive exPlanations)~\cite{Lundberg:2017:UAIM} is a popular model-agnostic method for explaining ML predictions across diverse data types, such as tabular, image, and time-series data. SHAP determines the contribution of input features to a model's output across all possible feature combinations, known as coalitions. This is done by comparing the model output when a feature is included or excluded across all possible coalitions. Coalitions are constructed by masking, i.e., removing, features which are not included. The removal of features is performed by replacing them with non-informative replacement data.

SHAP's computational cost grows exponentially with the number of input features. Therefore, approximation methods such as KernelSHAP were proposed~\cite{Lundberg:2017:UAIM}. However, KernelSHAP does not inherently account for dependencies in time-series data, posing a problem for CPS use-cases, which commonly include time-series data. To overcome this issue, specialised methods such as TimeSHAP~\cite{Bento:2021:TimeSHAP} and WindowSHAP~\cite{Nayebi:2023:WindowSHAP} were introduced to both accelerate SHAP calculations and incorporate temporal dependencies. 

C-SHAP~\cite{Jutte:2025:CTSA} offers another approach, whereas TimeSHAP and WindowSHAP determine the contribution of data points and segments, C-SHAP considers the contribution of global concepts. Concepts can be defined as high-level (as opposed to low-level) input features~\cite{Goyal:2020:CaCE}. For example, Sun \textit{et al.}~\cite{Sun:2023:EACS} define image segments as concepts and Kim \textit{et al.}~\cite{Kim:2018:TCAV} allow any kind of visual observation such as \enquote*{stripes} or \enquote*{female}. C-SHAP~\cite{Jutte:2025:CTSA} proposes to use components from signal decomposition as concepts for time-series data.

\subsection{Signal decomposition}
Signal decomposition is a well-established technique, applicable to time-series data, with examples such as: Discrete Wavelet Transform (DWT)~\cite{Heil:1989:CDWT}, Empirical Mode Decomposition (EMD)~\cite{Huang:1998:EMDH}, and Short-Time Fourier Transform (STFT)~\cite{Griffin:1984:SEMS}. While these techniques yield mathematically meaningful components, their practical interpretation can be challenging.

The custom decomposition approach by~\cite{Jutte:2025:CTSA} is intended for human interpretability. Instead of considering how a signal can be mathematically decomposed, the emphasis is on considering which concepts within the data are most relevant in understanding its behaviour. Examples of such concepts include classical signal components such as \emph{Bias} (the mean level), \emph{Trend} (the global change), and \emph{Variance}, representing changes in amplitude over time.

% ===============================================
% Section
% ===============================================
\section{Model improvement}
\label{sec:model_improvement}

\subsection{Use-case}
\label{subsec:use_case}
\begin{figure}[htbp]
	\centering
	\includegraphics[width=\linewidth]{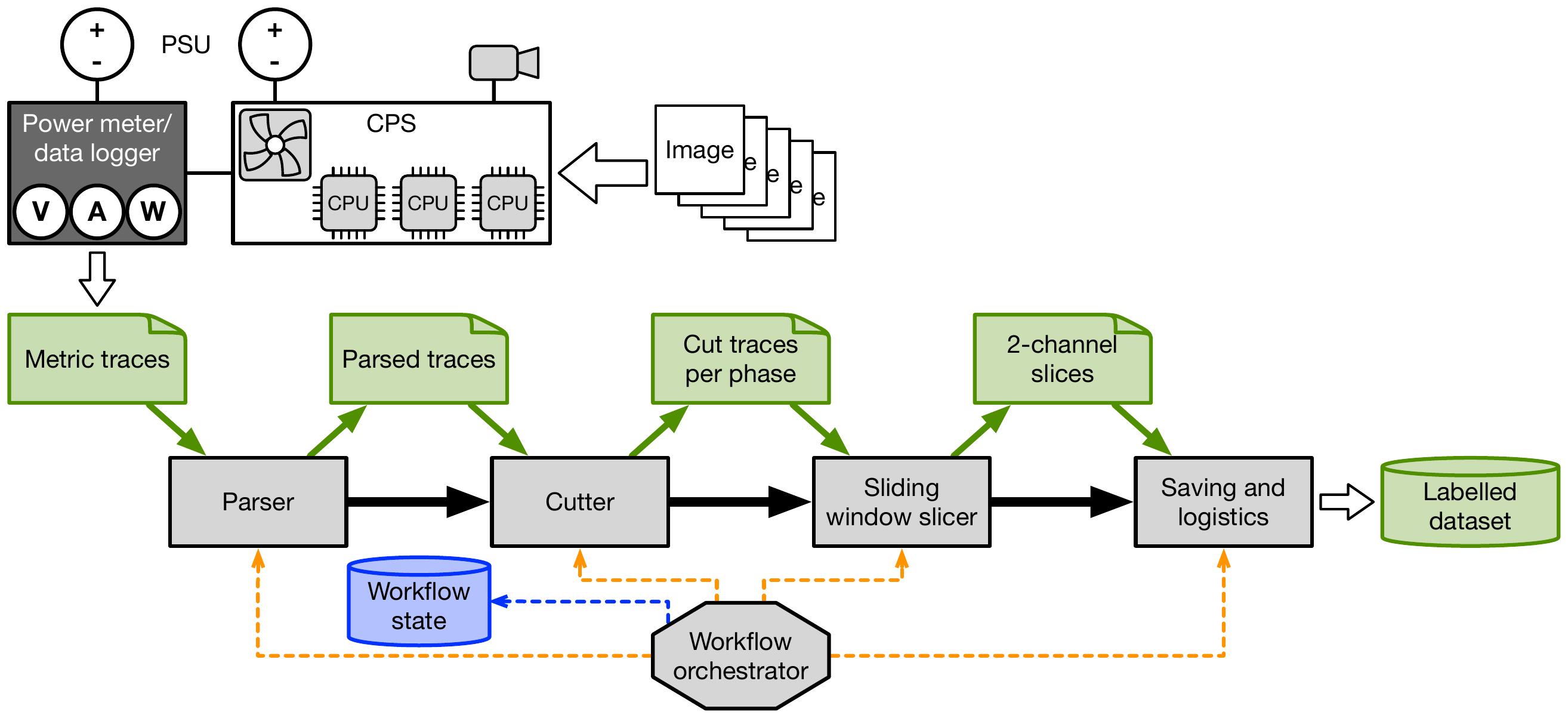}
	\caption{The experimental setup for machine trace collection and the data processing steps leading to a labelled dataset, suitable for training a CNN model.}
	\label{fig:ml_wf_cnn_dataset}
\end{figure}

We apply the SHAP-based explainability approach to an anomaly detection and identification solution~\cite{Odyurt:2021:PPFT}, with the aim of improving its performance. For the use-case, time-series traces collected from an experimental platform are processed to identify operational anomalies. This data-centric solution relies on Convolutional Neural Networks (CNNs), trained in a supervised fashion. The considered platform execution conditions (classes) are \emph{Normal}, \emph{NoFan}, and \emph{UnderVolt}, with the last two being anomalous. Electrical metrics such as current, voltage, and energy are monitored, with \emph{current} being the most effective. The data collection and processing steps are given in \Cref{fig:ml_wf_cnn_dataset}.

\paragraph*{Experimental setup}
The experimental platform consists of the ODROID-XU4\footnote{\url{https://www.odroid.nl/odroid-xu4}} computing device, monitored by a high-frequency power data logger. To achieve highest isolation, the computing device and the power logger are attached to individual Power Supply Units (PSUs). The processor design follows the ARM big.LITTLE architecture. The computational workload, executed per input, is an object detection task, processing images provided through a camera or a storage medium. Power metric readings are collected under the aforementioned execution conditions. The NoFan execution represents operating with a faulty cooling fan, while the UnderVolt describes operating with inconsistent and under-provisioned voltage supply, both affecting the performance and stability of the platform. With different combinations of workload, processor core, and execution condition, 24 scenarios are defined. Experiments are not synthetic, nor simulated.

\paragraph*{ML workflow}
There are two ML workflows for this use-case, one to prepare a labelled dataset (\Cref{fig:ml_wf_cnn_dataset}) and another to train a CNN model. The workflow to train the CNN model is rather straightforward. Dataset formation in the former workflow relies on the compartmentalisation of data per execution phases, as explained in \Cref{sec:background}. The raw monitoring data collected in a continuous fashion during any particular execution is parsed and then cut to system-specific phase structures. The phase type intended to be used for training and inference shall be kept. We consider the full processing activity per image, i.e., \texttt{cycle-op} phases. As the full image processing phase contains more diverse behaviour, it results in a better demonstration of generating per concept scoring. A windowing algorithm is used to generate slices of data (instances) from compartmentalised phase data, alongside the relevant label.

\paragraph*{Dataset}
The resulting dataset in this case is well-balanced. Machine trace data is collected per scenario followed by a sliding window algorithm. Resulting data instances cover two dimensions, the temporal dimension (timestamps) and one metric dimension, i.e., electrical current values. Brief statistics for variations of this dataset with different sliding window parameters are listed in \Cref{tab:dataset_stats}. Instance count variation is the result of image processing time differences, i.e., execution time, which in turn is driven by the employed CPU core type.
\begin{table}[htbp]
    \centering
    \caption{Instance counts per sliding window profile for the \emph{training dataset}.}
    \label{tab:dataset_stats}
    \begin{tabular}{@{}lrrr@{}}
        \toprule
        \textbf{Statistic} & \textbf{Profile 100} & \textbf{Profile 200} & \textbf{Profile 400} \\
        \midrule
        Window size                     & 100           & 200           & 400 \\
        Window shift                    & 10            & 10            & 10 \\
        Per scenario instance (Max)     & 365\,136      & 362\,136      & 356\,136 \\
        Per scenario instance (Min)     & 12\,749       & 12\,449       & 11\,849 \\
        Per scenario instance (Mean)    & 136\,620      & 134\,973      & 131\,678 \\
        Total instances                 & 3\,278\,899   & 3\,239\,359   & 3\,160\,279 \\
        \bottomrule
    \end{tabular}
\end{table}

\paragraph*{ML model}
A CNN model is used for the scoring method. This classifier CNN network takes two data channels as input, the time channel and the metric channel. There are five sequential convolution layers with different output channel sizes of $[64, 64, 128, 128, 256, 256]$, all having kernel sizes of 5. The convolution block is followed by a single fully-connected layer of the size 4096. The predicted class is one of three aforementioned execution conditions.

\subsection{High-level approach}
\label{sec:high_level_approach}
\begin{figure}[htbp]
	\centering
    \includegraphics[width=\linewidth]{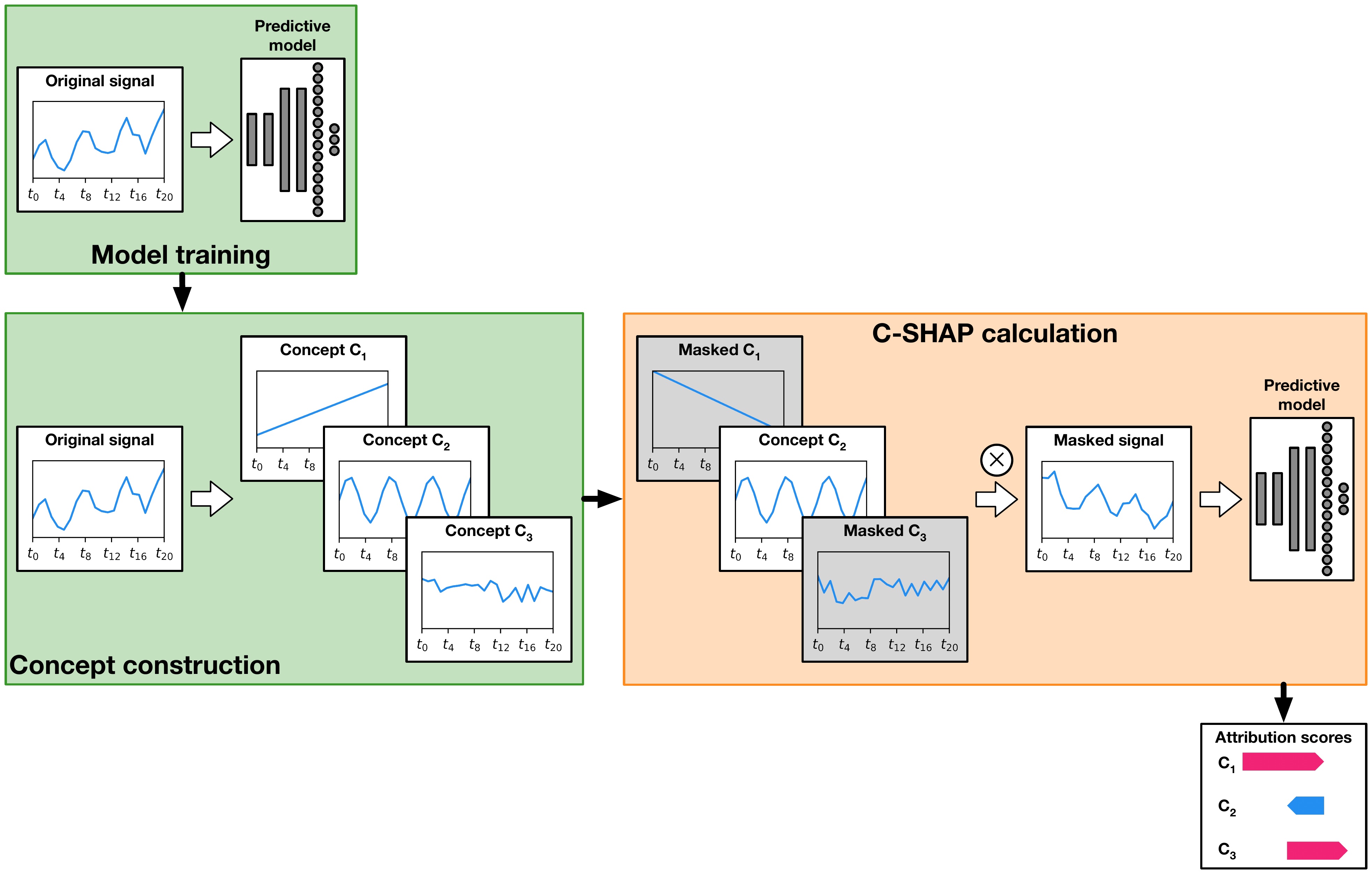}
	\caption{C-SHAP workflow, including concept construction from signal data, masking and calculation of C-SHAP, i.e., attribution scores.}
	\label{fig:cshap_wf}
\end{figure}

The C-SHAP workflow for post-hoc explanations consists of three steps: model training, concept construction and calculation of the SHAP values, as depicted in \Cref{fig:cshap_wf}. We already covered the model training. What remains is the concept construction using time-series decomposition, as will be discussed next, and the calculation of the SHAP values.

For the concept construction, time-series decomposition is used. Existing decomposition methods come with limitations, hindering their applicability to our use-case data. For example, EMD cannot represent the discrete jumps present in our data. Initial experiments with the DWT revealed its capability for dealing with jumps using the Haar wavelet. However, the interpretability of the resulting components is questionable. For C-SHAP to be useful, domain experts need to be able to understand the meaning of considered components. We use a custom decomposition designed specifically for interpretability, as discussed in \Cref{sec:background}. The defined concepts are: \enquote*{Levels}, \enquote*{Peaks}, \enquote*{Scale}, \enquote*{Low Frequency (LF)} and \enquote*{High Frequency (HF)}. We model the \enquote*{Scale} component as multiplicative, while the other components are additive. This results in the decomposition:
\begin{equation}
     \mathbf{y}(t) = \mathbf{y}_\text{Levels}(t) + \mathbf{y}_\text{Peaks}(t) + \mathbf{y}_\text{Scale}(t) \times (\mathbf{y}_\text{LF}(t) + \mathbf{y}_\text{HF}(t)).
\label{eq:custom_decomposition}
\end{equation}

% ===============================================
% Section
% ===============================================
\section{Implementation}
\label{sec:implementation}

\subsection{Concept construction}
To generate the C-SHAP, i.e., SHAP scores, to determine concept contribution in model prediction, concepts are constructed from the input data. The concept construction uses the decomposition defined in \Cref{eq:custom_decomposition}. For the implementation of this decomposition, components are extracted from left to right.

First, the \enquote*{Levels} component is constructed, modelling the local mean of the signal. We achieve this using Change Point Detection (CPD), which detects changes in the data distribution, yielding segments with similar behaviour. The \enquote*{Levels} component is modelled as the mean of the signal on these segments. Change points are detected using the Pruned Exact Linear Time (PELT) algorithm~\cite{Killick:2012:ODCL}. For the detection, a Radial Basis Function (RBF) cost function is used, with a subsampling of 40 points and a penalty of 50. Residuals resulting from removing the level changes are filtered from the signal. Specifically, points within 20 time-steps of a level change are resampled from a normal distribution with the mean and standard deviation of the preceding segment. The resampled values are smoothened using a moving average with window size 20.

Next, the \enquote*{Peaks} component is modelled as statistical outliers, i.e., points with values below the first quantile or above the third quantile. To maintain plausibility of the filtered signal, peaks are not replaced by zero, but replaced by the \emph{median of the signal}. Noise is added to the replacement values with zero mean and a standard deviation equal, matching that of the peak-filtered signal. To maintain the smoothness of the signal, a moving average is applied to the noise before adding it to the signal.

The \enquote*{Scale} component is modelled as the maximum absolute value of the remaining signal. The \enquote*{Low frequency} component is calculated as the moving average of the signal with a window of 75 points. Finally, the \enquote*{High frequency} component represents the residual of the signal.

\subsection{C-SHAP calculation}
The SHAP values are calculated exactly, rather than using approximation methods such as KernelSHAP~\cite{Lundberg:2017:UAIM}. To mask features, i.e., concepts, with uninformative data they are replaced by concepts sampled from the training data, a common practice in literature. We use training data, rather than zero values, which are unrepresentative of our data distribution (a sample with no high-frequency behaviour is unrealistic) and can lead to out-of-distribution effects inherent in SHAP-based methods.

This training data is selected from scenarios corresponding to the same CPU core type and number of execution rounds. This results in a selection of six scenarios (out of 24). From each scenario we randomly select five cycles. The choice of five was made to capture diversity while limiting computational costs. To allow for the substitution of components, the selected samples need to match the length of the original components. Longer samples are truncated by removing data from the end, shorter samples are padded by repeating the final value.

\subsection{Test data}
C-SHAP is applied to data instances kept aside during training, to be used for final inference. We select two samples (\texttt{cycle-op} phases) from 3 scenarios, each belonging to one of the execution classes, i.e., Normal, NoFan, and UnderVolt. We select the samples from early and late points in the execution timeline to reflect the variations and effects of anomalies. The instances generated by running the windowing algorithm over these samples forms our test data.

% ===============================================
% Section
% ===============================================
\section{Outcome}
\label{sec:outcome}
\begin{figure*}[htbp]
    \begin{minipage}[c]{0.49\textwidth}
        \centering
        \includegraphics[width=\linewidth]{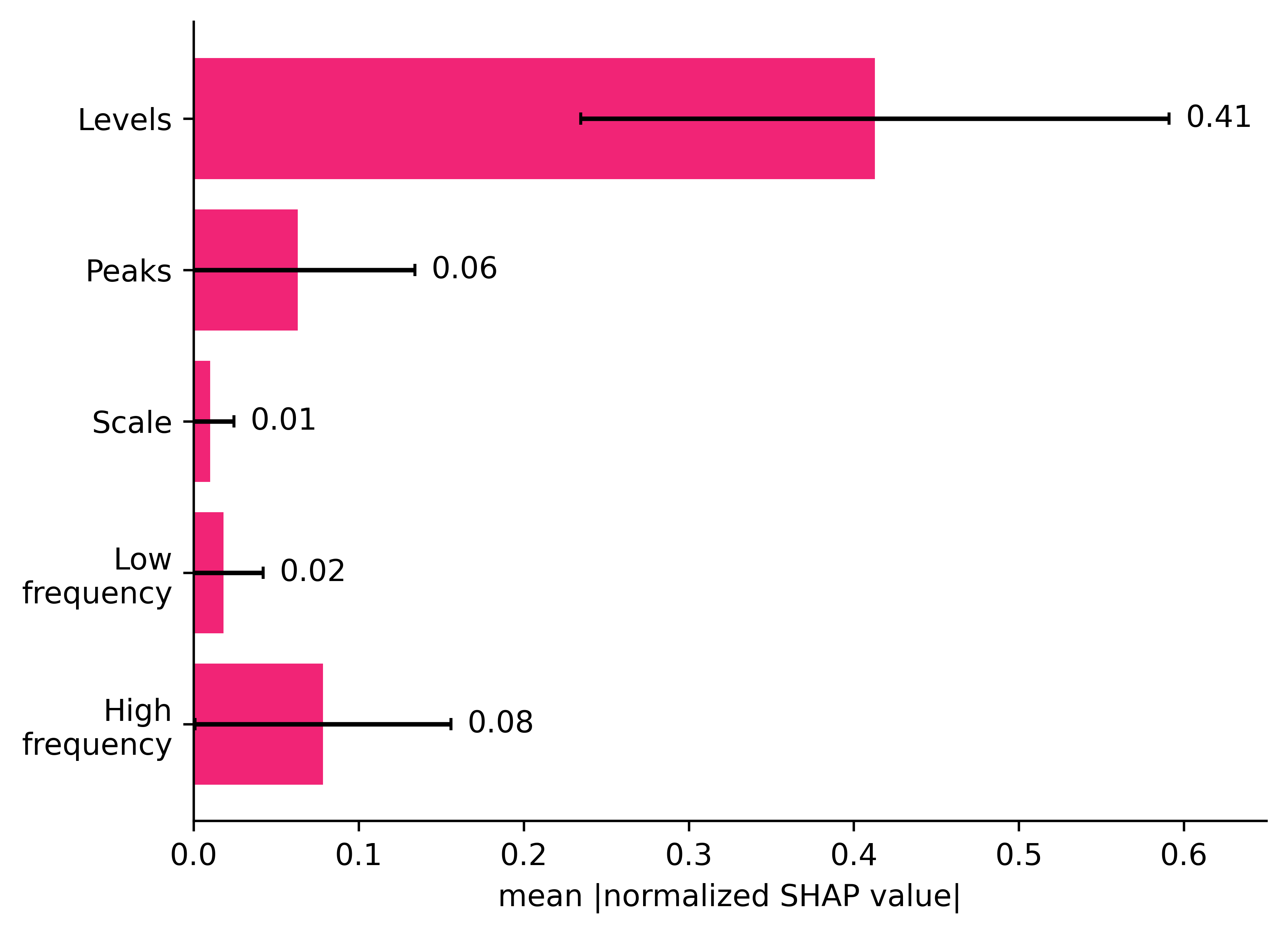}
        \caption{The mean absolute SHAP values (with respect to the ground truth classes) over the test data with respect to the concepts.}
        \label{fig:glob_exp}
    \end{minipage}
    \hfill
    \begin{minipage}[c]{0.49\textwidth}
        \centering
        \includegraphics[width=\linewidth]{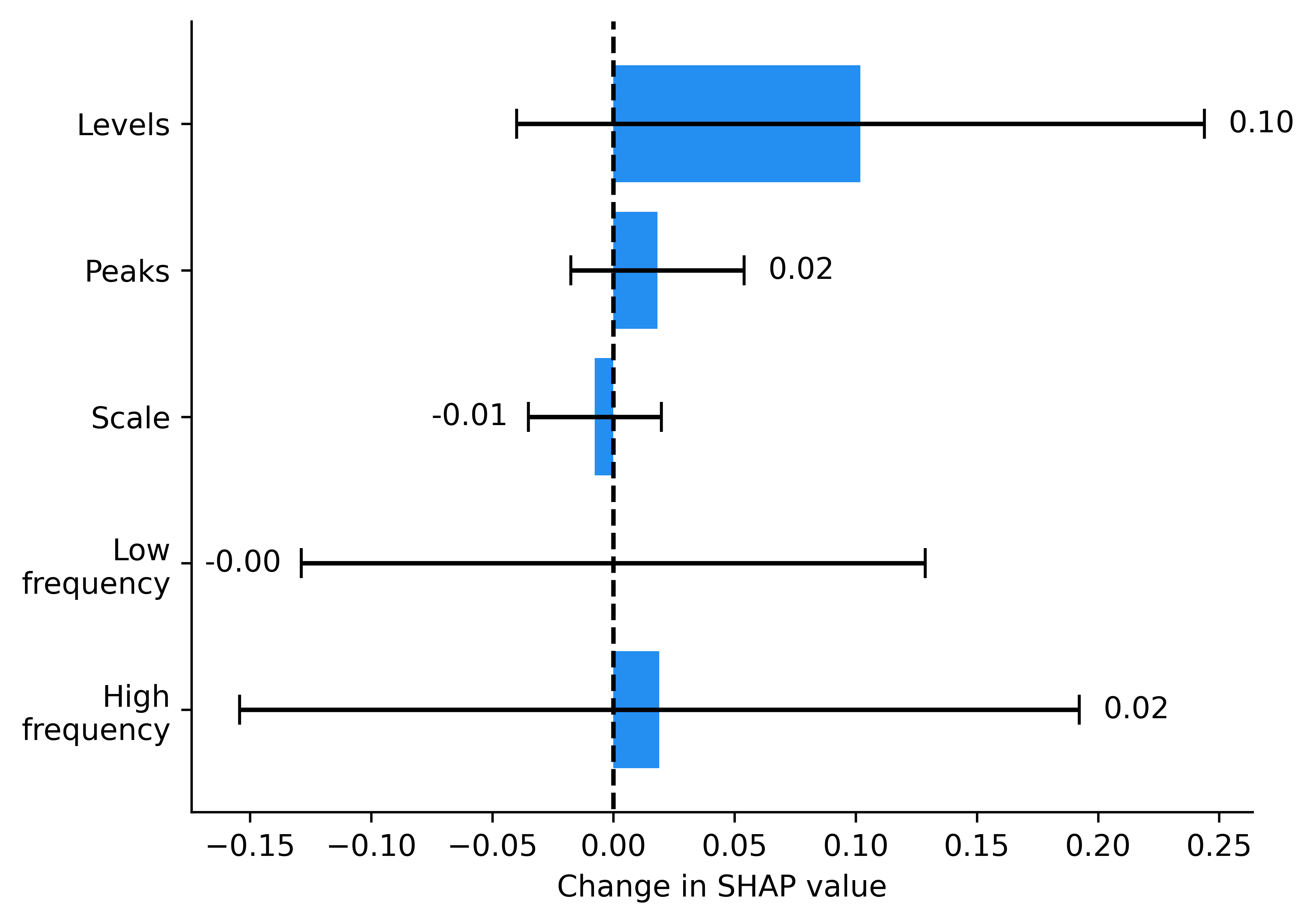}
        \caption{Change in SHAP values (mean and standard deviation) for the test data when increasing the window size from 100 to 400.}
        \label{fig:change_shap_100_400}
    \end{minipage}
\end{figure*}

The CNN model trained on windows of size 100 achieved an accuracy of 83.78\% on the test data. \Cref{fig:glob_exp} shows the mean absolute C-SHAP values over all test data. The \enquote*{Levels} concept is the most influential on the model's predictions, followed by \enquote*{High frequency} and \enquote*{Peaks}. Hence, the model mainly relies on the local level of the current, while also considering rapid fluctuations. The scale of the fluctuations and slow changes over time are considered less important.

\begin{figure}[htbp]
    \centering
    % Top row
    \begin{subfigure}[b]{0.49\textwidth}
        \setcounter{subfigure}{0}
        \includegraphics[width=\linewidth]{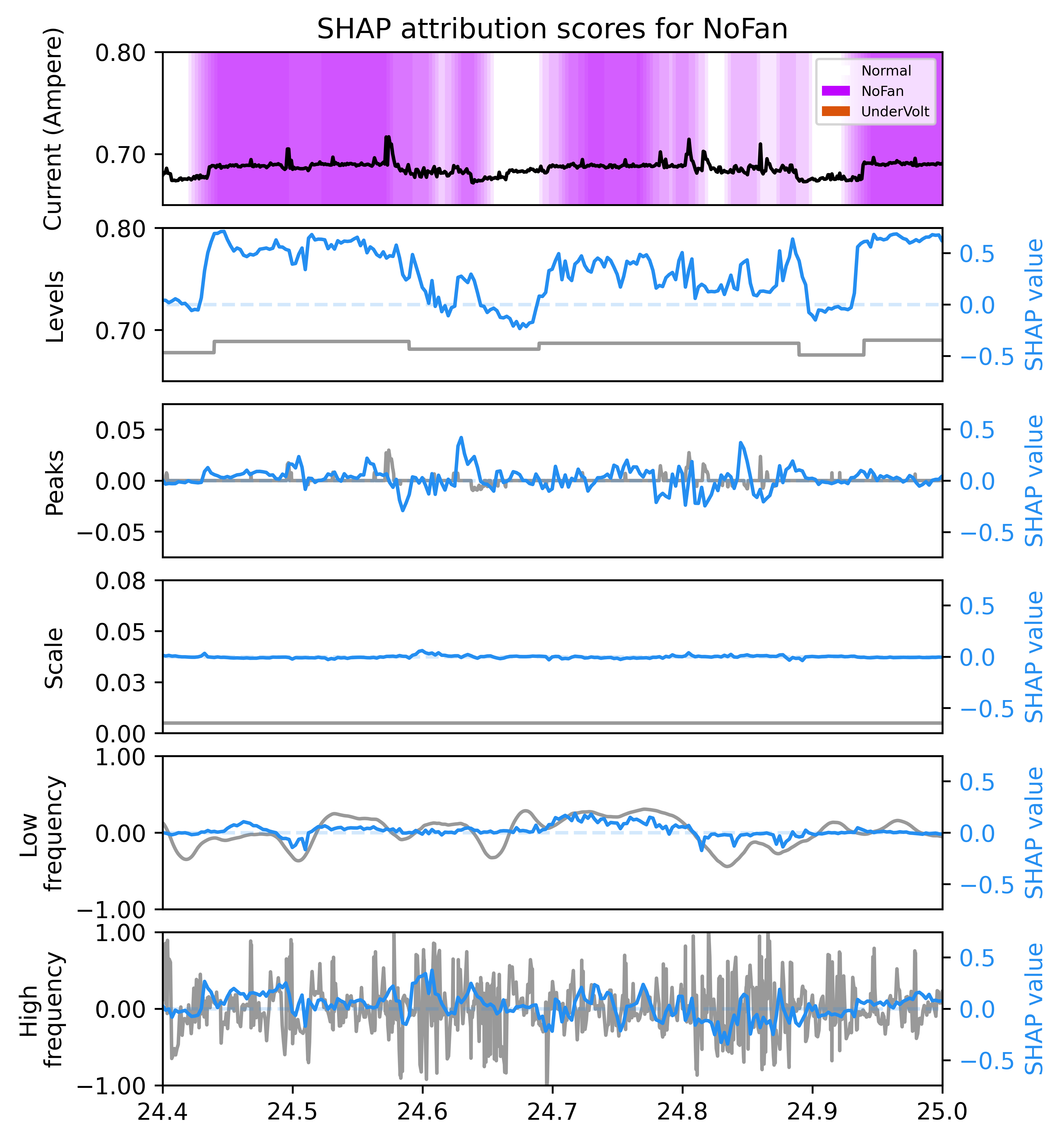}
        \caption{Example NoFan, window size 100.}
        \label{fig:nofan_100}
    \end{subfigure}
    \hfill
    \begin{subfigure}[b]{0.49\textwidth}
        \includegraphics[width=\linewidth]{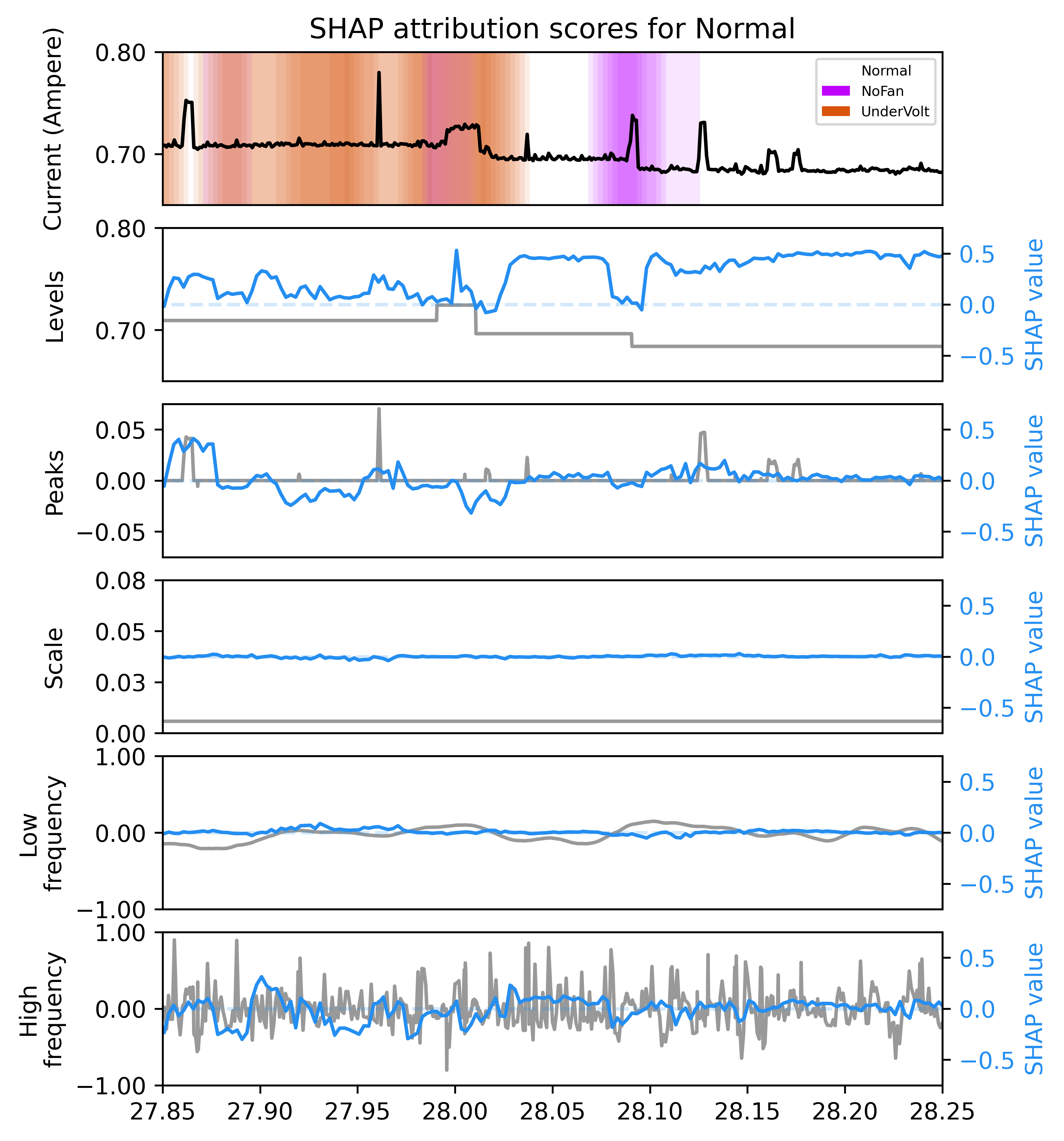}
        \caption{Example Normal, window size 100.}
        \label{fig:normal_100}
    \end{subfigure}
    % Middle row
    \begin{subfigure}[b]{0.49\textwidth}
        \includegraphics[width=\linewidth]{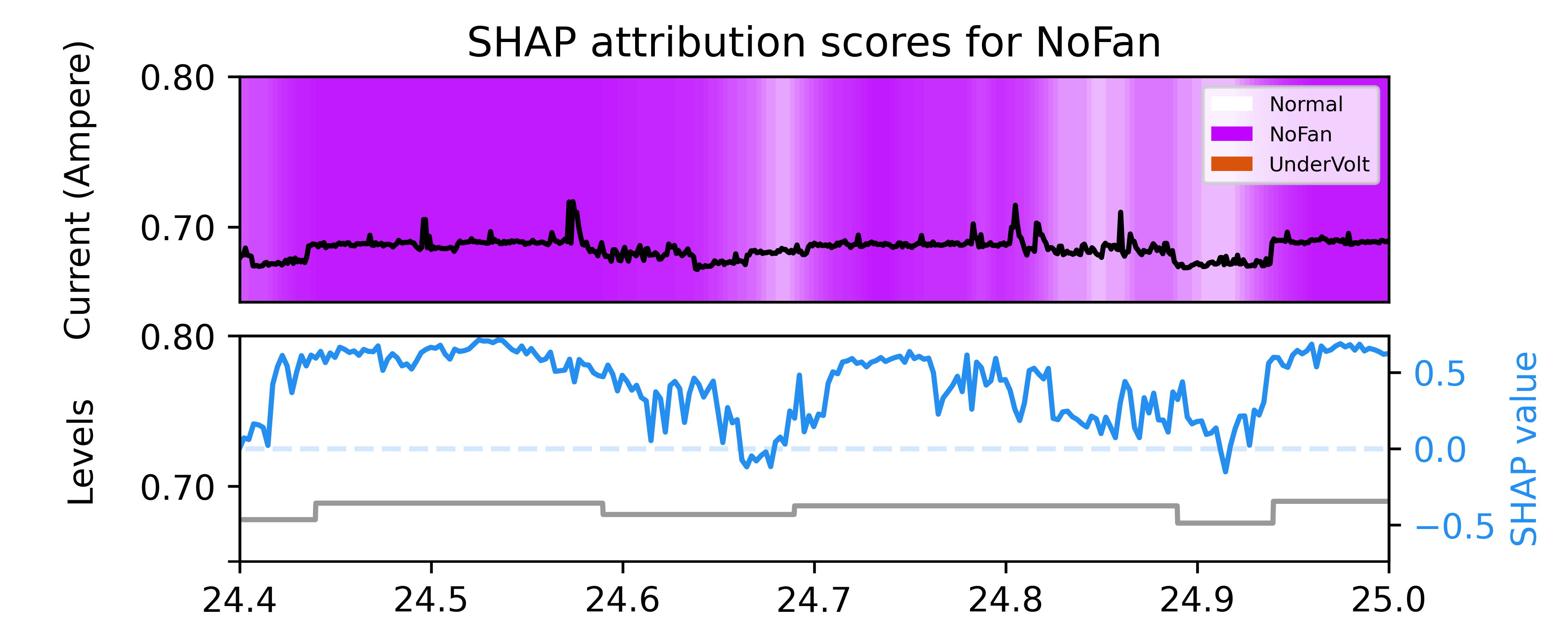}
        \caption{Example NoFan, window size 200.}
        \label{fig:nofan_200}
    \end{subfigure}
    \hfill
    \begin{subfigure}[b]{0.49\textwidth}
        \includegraphics[width=\linewidth]{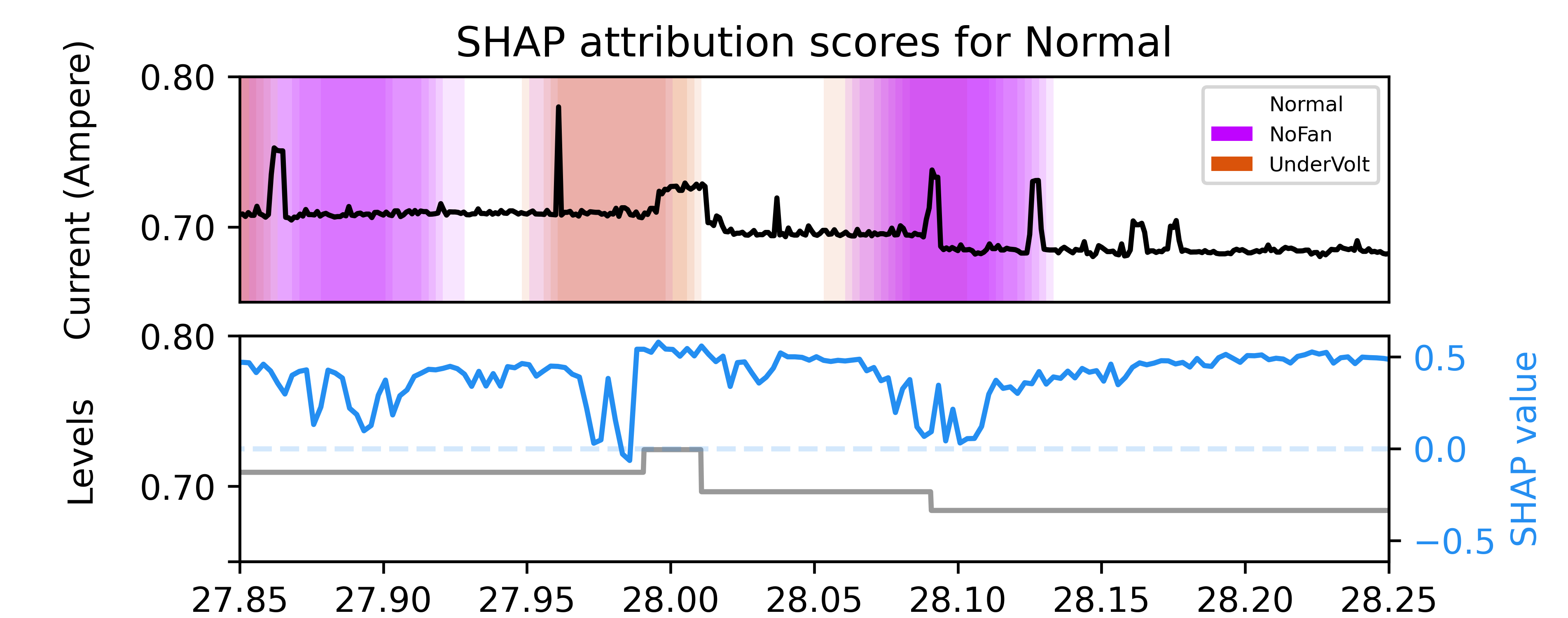}
        \caption{Example Normal, window size 200.}
        \label{fig:normal_200}
    \end{subfigure}
    % Bottom row
    \begin{subfigure}[b]{0.49\textwidth}
        \includegraphics[width=\linewidth]{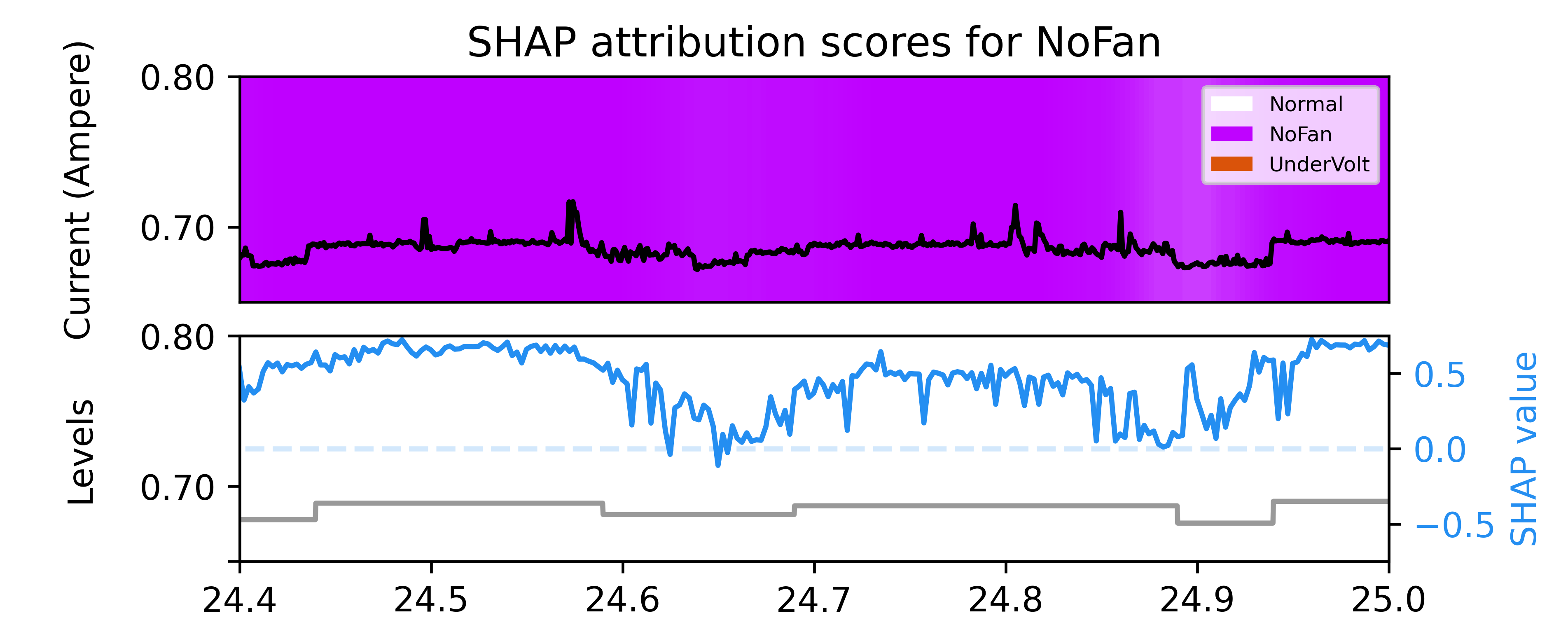}
        \caption{Example NoFan, window size 400.}
        \label{fig:nofan_400}
    \end{subfigure}
    \hfill
    \begin{subfigure}[b]{0.49\textwidth}
        \includegraphics[width=\linewidth]{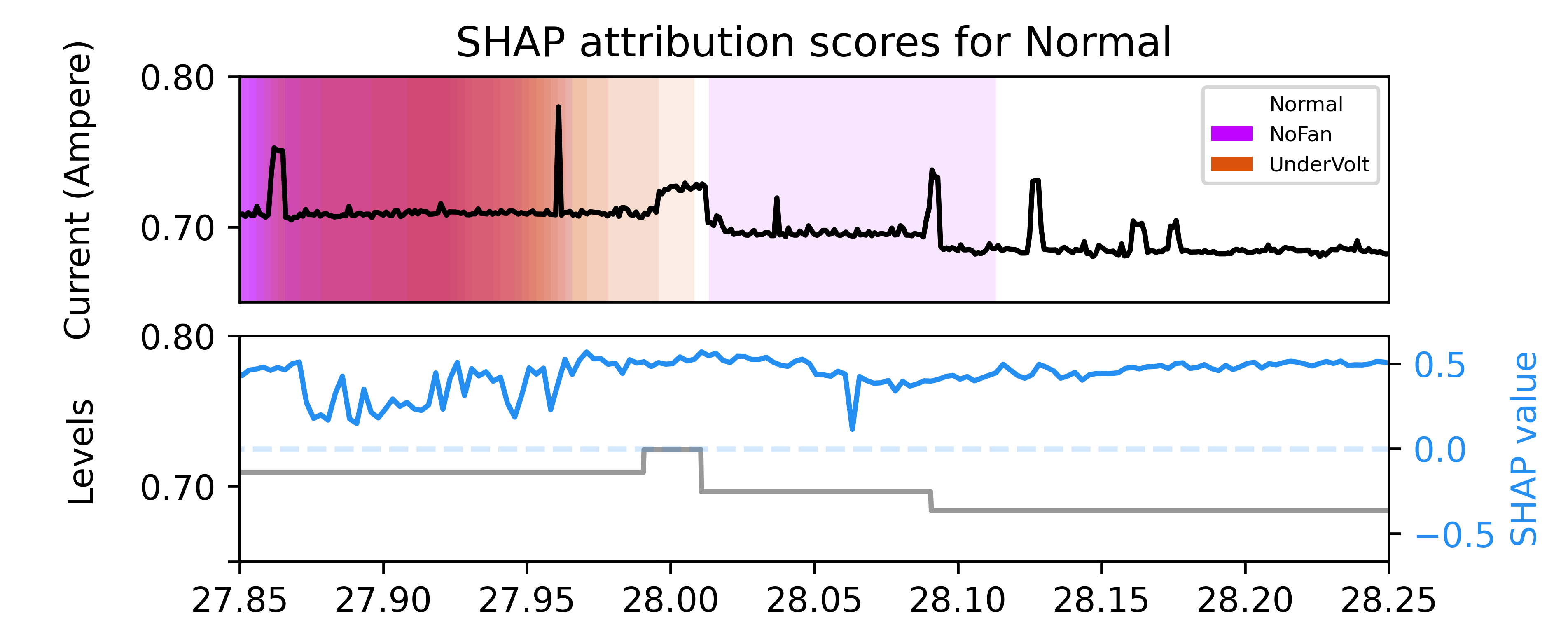}
        \caption{Example Normal, window size 400.}
        \label{fig:normal_400}
    \end{subfigure}
    \caption{The predictions and SHAP values for two selected example segments from the inference data. a), c) and e) correspond to segments of the same NoFan sample, while b), d) and f) correspond to segments of the same Normal sample. a) and b) correspond to the results for window size 100, while c) and d) correspond to size 200, and e) and f) correspond to 400. In the top rows of the images, the signal is shown along with the predicted classes as coloured overlays, marking the associated windows. For the NoFan sample, the purple windows are correct, and for the Normal sample, the white windows are correct. For each component, the component itself is plotted, combined with the SHAP values for each window. For c)-f) only the \enquote*{Levels} component is shown for clarity.}
    \label{fig:local_exp}
\end{figure}

These results are illustrated in \Cref{fig:nofan_100,fig:normal_100}, visualising two segments of test samples with the SHAP values for the concepts of the associated windows. \enquote*{Levels} generally displays higher SHAP values than the other concepts. Importantly, drops in the \enquote*{Levels} SHAP values coincide with misclassifications, demonstrating the influence of the \enquote*{Levels} concept on these errors.

\begin{figure}[htbp]
    \centering
    \begin{subfigure}[t]{0.49\textwidth}
        \centering
        \includegraphics[width=\linewidth]{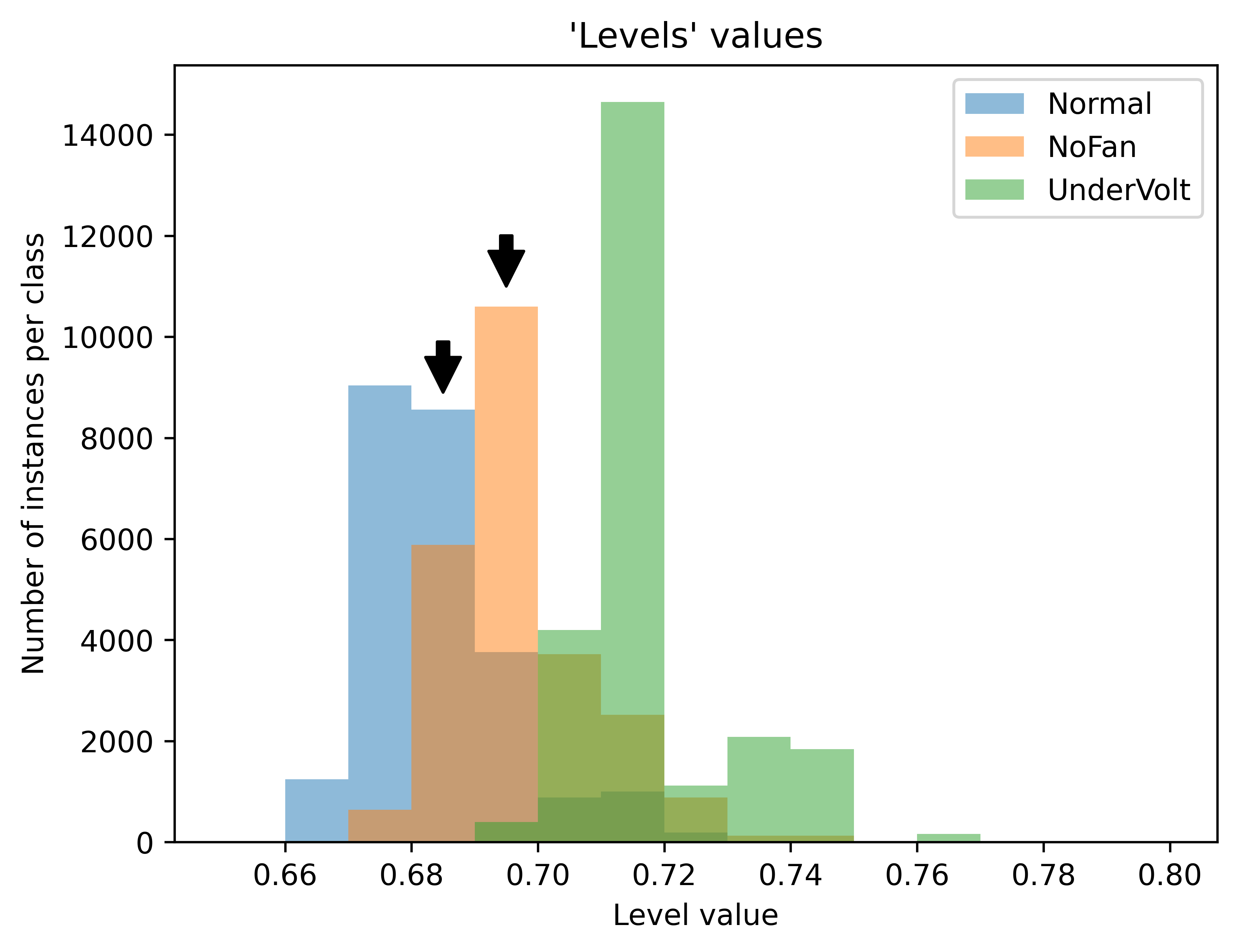}
        \caption{Training data.}
        \label{fig:hist_train}
    \end{subfigure}
    \hfill
    \begin{subfigure}[t]{0.49\textwidth}
        \centering
        \includegraphics[width=\linewidth]{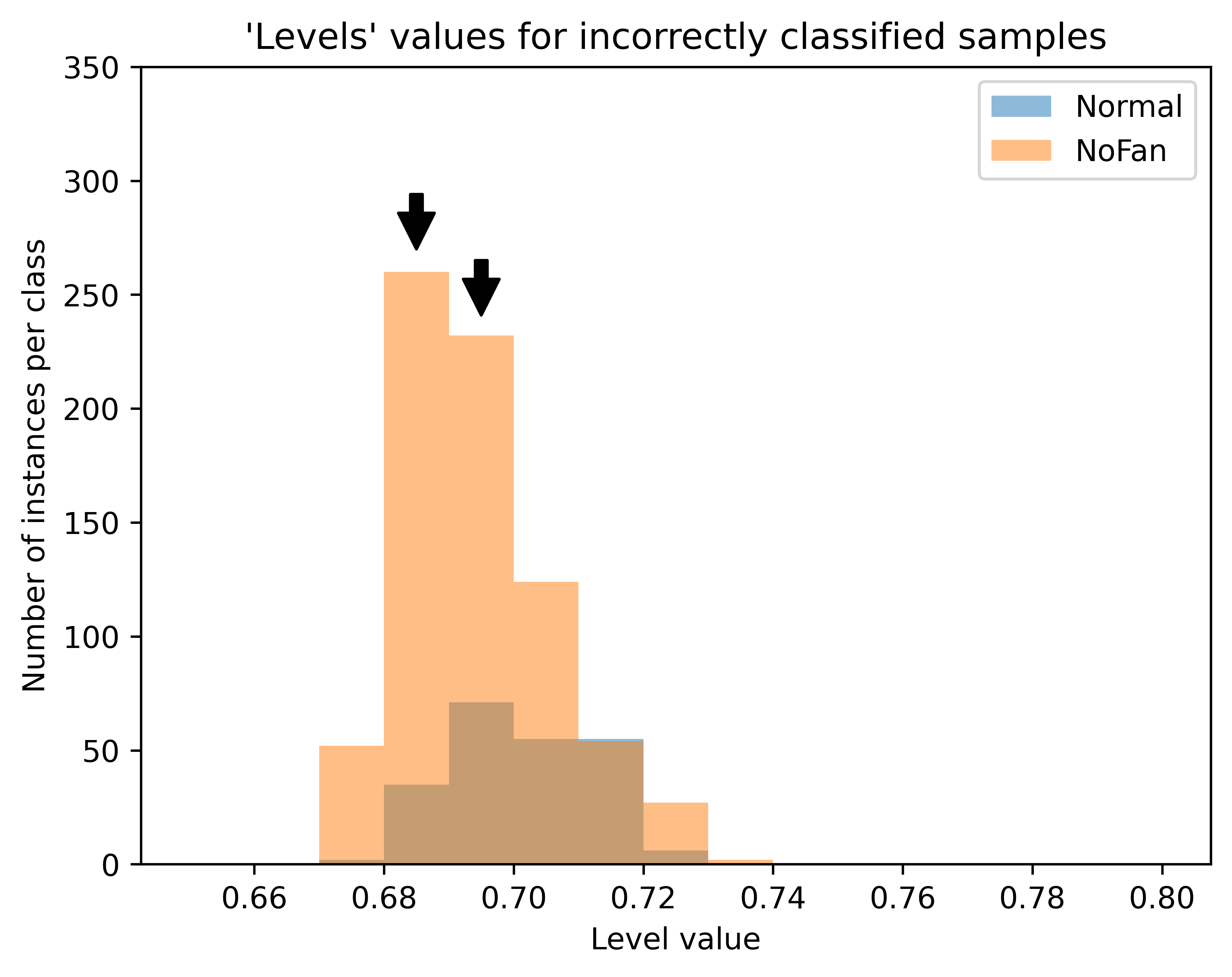}
        \caption{Misclassified inference data.}
        \label{fig:hist_misclassified}
    \end{subfigure}
    \caption{Value histograms for component \enquote*{Levels}, per class label. The arrows in both figures mark the bins with most misclassifications for NoFan.}
    \label{fig:hist_level_values}
\end{figure}

The histogram of \enquote*{Levels} values in the training data (\Cref{fig:hist_train}) shows there is a class separation based on level, but also an overlap between classes. Analysis of misclassified samples of the classes Normal and NoFan (\Cref{fig:hist_misclassified}) reveals that many errors occur in the overlapping \enquote*{Levels} value range. We focus on these two classes, since we found that most errors occur due to the confusion between them. The confusion is specifically apparent between 0.680--0.685, where Normal is prevalent, and 0.685--0.700, where NoFan is prevalent.

We hypothesised that increasing the window size would provide more contextual information and reduce the ambiguity related to the \enquote*{Levels} values, since more windows may include multiple levels. Training models with window sizes of 200 and 400 data points, improved test accuracies to 87.9\% and 92.3\% respectively. Comparing SHAP values between the models with window sizes 100 and 400 (\Cref{fig:change_shap_100_400}), illustrates that this improvement is matched by an increase in the SHAP values for the \enquote*{Levels} concept, as hypothesized. \Cref{fig:local_exp} shows that the new models seem to display more consistent SHAP values for \enquote*{Levels}. This increased stability is confirmed by the decrease in standard deviation of the mean absolute SHAP value of the \enquote*{Levels} concept from 0.178 for window size 100, to 0.170 for window size 200, and to 0.155 for window size 400.

% ===============================================
% Section
% ===============================================
\section{Related work}
\label{sec:related_work}
The use of SHAP values for feature selection has been covered in the literature, e.g.,~\cite{Lundberg:2017:UAIM,Wang:2024:FSSC}. While for traditional models, the effect of high-level features is direct~\cite{Khan:2022:ERFP}, the control over features for deep learning models such as CNNs is generally indirect. These models extract suitable high-level features themselves, which can only be influenced through hyperparameter or data manipulation. Common examples of XAI through the SHAP scoring of high-level features revolve around image analysis use-cases~\cite{Kawauchi:2022:SIOD,Dardouillet:2023:EISS}. As per context of our use-case, we focus on solutions for industrial CPS and apply SHAP scoring to features extracted from time-series data, which portray complex behaviour.

Within the literature, while claiming application of XAI in solutions intended for CPS domain, shortcomings are apparent. Not every considered use-case can be designated as a CPS in its industrial sense, e.g., a hard disk drive~\cite{Ferraro:2023:EEAI}. Such loose categorisations are seen in CPS-focused surveys~\cite{Hoenig:2024:EACP}. The main cause is perhaps the variable definition of the term \enquote*{CPS} itself.

% ===============================================
% Section
% ===============================================
\section{Conclusion and future work}
\label{sec:conclusion}
We described an approach to arrive at effective ML model hyperparameter adjustments, relying on XAI techniques. We argued the advantage over design-space search methodologies, as XAI provides the designer with reasoning behind the adjustment, making the model superior in terms of reliability and dealing with future, unseen data. We have shown how this approach is implemented for time-series data collected from industrial CPS machine operations, by considering custom, human-interpretable signal decompositions. The improvements in prediction performance is demonstrated using our experimental platform. It must be noted that while our method of applying XAI to uncover model reasoning is generalisable, both the model reasoning itself and the potential improvements presented in this paper are dataset and model specific.

%There can be other implications beyond the ML model itself.
Given the component scoring, upon consistent reliance on a particular component, a less complex architecture could be considered, or the opposite could be argued. Plus, depending on component importance, a lower sample rate could be argued in favour of. This means that the monitoring subsystem can be adjusted accordingly, leading to lower data sampling overhead and data size reduction. Additionally, the measures taken based on the XAI findings should be compared with the other techniques, such as automated hyperparameter tuning and alterative XAI approaches, e.g., WindowSHAP~\cite{Nayebi:2023:WindowSHAP}. These can be interesting future work avenues to explore.

% ===============================================
% Acknowledgements
% ===============================================
\begin{credits}
\subsubsection{\ackname} This publication is part of the project ZORRO with project number KICH1.ST02.21.003 of the research programme Key Enabling Technologies (KIC), which is (partly) financed by the Dutch Research Council (NWO).
\end{credits}

% \begin{credits}
% \subsubsection{\ackname} Redacted ...
% \end{credits}

% ###############################################
% End of file
% ###############################################

% ===============================================
% Balance columns for the last page
% ===============================================
\balance

% ###############################################
% Bibliography
% ###############################################
\bibliographystyle{splncs04}
\bibliography{bibliography/references.bib}

% ###############################################
% Document end
% ###############################################
\end{document}